%% file: root.tex
\documentclass[letterpaper, 10 pt, conference]{ieeeconf}  
\IEEEoverridecommandlockouts                              
\pdfoutput=1
\usepackage{hyperref}
\usepackage{amssymb}  
\usepackage{algorithm} 
\usepackage{algorithmic}
\usepackage{graphicx}
\graphicspath{ {icra2019/figs/} }
\usepackage{float}

\usepackage{
    amsmath,
    float,
    xcolor,
    marginnote,
    subcaption,
    pgfgantt,
    times,
    outlines,
    wrapfig,
    romannum,
    csquotes,
    bm,
}
\usepackage[doi=false,isbn=false,defernumbers=true,style=numeric,backend=bibtex,maxbibnames=1000,sorting=none,firstinits=true]{biblatex}

\title{\LARGE \bf
Reinforcement Learning on Variable Impedance Controller for High-Precision Robotic Assembly 
}

\author{ Jianlan Luo$^{1}$, Eugen Solowjow$^{2}$, Chengtao Wen$^{2}$, Juan Aparicio Ojea$^{2}$, Alice M. Agogino$^{1}$ \\Aviv Tamar$^{3}$, Pieter Abbeel$^{1}$
\thanks{$^{1}$University of California, Berkeley, CA 94704, USA}
\thanks{$^{2}$Siemens Corporate Technology, Berkeley, CA, 94704, USA}%
\thanks{$^{3}$Technion, Haifa, 3200003, Israel}
}

\begin{document}

\maketitle

\newcommand{\bx}{{\mathbf x}}
\newcommand{\bu}{{\mathbf u}}
\newcommand{\bK}{{\mathbf K}}
\newcommand{\bk}{{\mathbf k}}
\newcommand{\bC}{{\mathbf C}}
\newcommand{\bo}{{\mathbf o}}
\newcommand{\bF}{\mathcal{F}}
\newcommand{\btau}{{\mathbf {\tau}}}
\begin{abstract}
Precise robotic manipulation skills are desirable in many industrial settings, reinforcement learning (RL) methods hold the promise of acquiring these skills autonomously. In this paper, we explicitly consider incorporating operational space force/torque information into reinforcement learning; this is motivated by humans heuristically mapping perceived forces to control actions, which results in completing high-precision tasks in a fairly easy manner. Our approach combines RL with force/torque information by incorporating a proper operational space force controller; where we also exploit different ablations on processing this information.  Moreover, we propose a neural network architecture that generalizes to reasonable variations of the environment. We evaluate our method on the open-source Siemens Robot Learning Challenge, which requires precise and delicate force-controlled behavior to assemble a tight-fit gear wheel set.
\end{abstract}

\input{sections/01_introduction.tex}\label{introduction}
\input{sections/02_problem.tex}\label{problem}
\input{sections/03_prelims.tex}\label{prelims}
\input{sections/04_rl_force.tex}\label{rl_force}
\input{sections/05_experiments.tex}\label{experiment_sec}
\input{sections/06_conclusion.tex}
\input{sections/acknowledgement.tex}


\printbibliography
\end{document}

%% file: sections/01_introduction.tex
\section{INTRODUCTION}
Today, industrial robots deployed across various industries are mostly doing repetitive tasks. The overall task performance hinges on the accuracy of their controllers to track pre-defined trajectories. 
To this end, endowing these machines with a greater level of intelligence to autonomously acquire skills is desirable.
The main challenge is to design adaptable, yet robust, control algorithms in the face of inherent difficulties in modeling all possible system behaviors and the necessity of behavior generalization.
Reinforcement learning (RL) methods hold promises for solving such challenges, because they promise agents to learn behaviors through interaction with their surrounding environments and ideally generalize to new unseen scenarios \cite{sutton1998reinforcement,deisenroth2013survey,kober2013reinforcement,levine2016end}. 

\begin{figure}[t]
    \centering
   \includegraphics[width=0.45\textwidth,keepaspectratio]{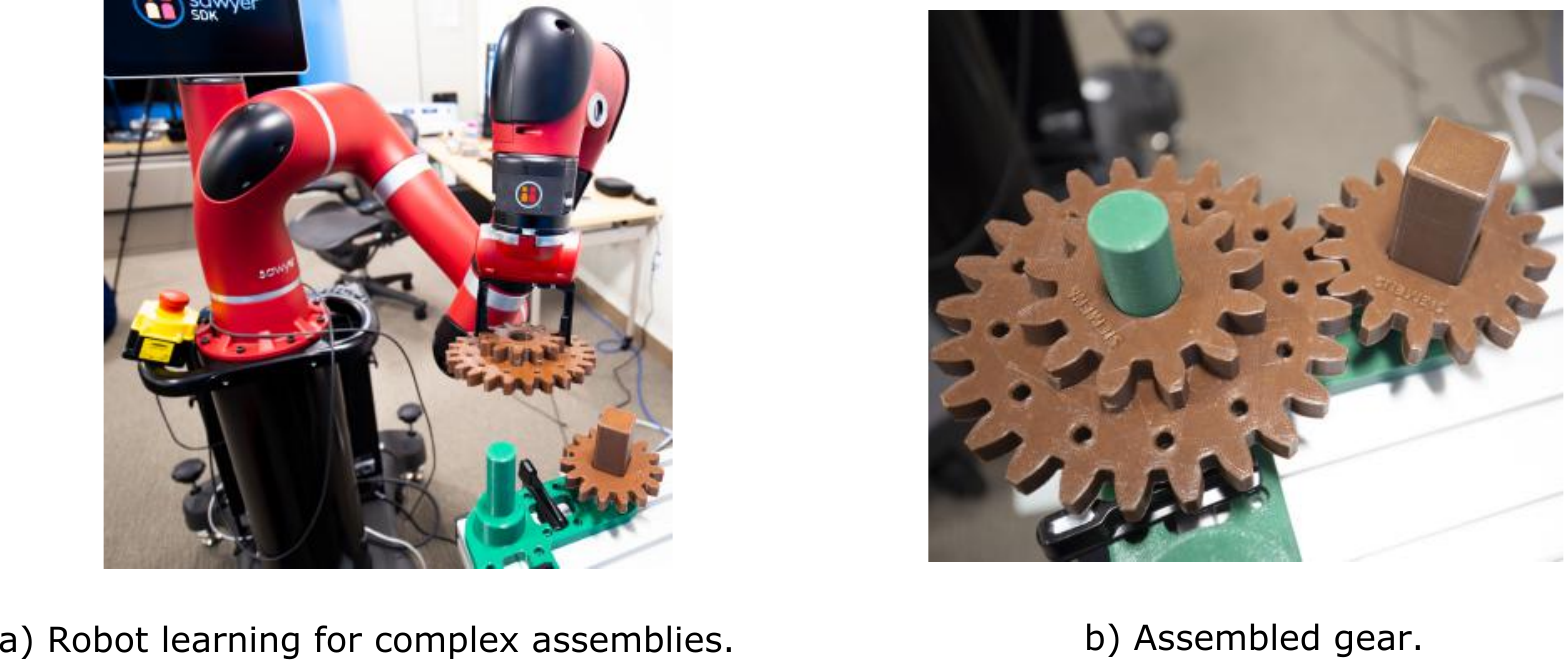}
    \caption{Learning control policies for assembly tasks.}
    \label{cover_photo}
\end{figure}

In this paper, we aim to learn policies that can assemble a high-precision gear set as shown in Fig.\ref{cover_photo}. In real manufacturing, human labor can accomplish such high-accuracy complex tasks in a fairly easy manner. For example, a peg in hole insertion is achieved by ``feeling" the contacts. This can be achieve with heuristics based on force feedback, for instance by probing the hole before inserting or moving the peg around the surface to search for the insertion point. However, designing robust strategies by properly processing observations is more desirable than heuristics or estimating perfect physical dynamics. RL allows to find control policies automatically for problems where traditionally heuristicis have been used. The question arises \textit{how do we properly integrate observed force information into reinforcement learning process to produce desirable behaviors?}

 RL is a method for learning such reactive policies automatically, through trial and error interaction in the domain, guided only by a reward signal that specifies how well the robot is performing the task. In practice, RL requires an informative reward signal to works effectively, which can be hard to design automatically. With sparse reward that just specifies successful task completion, RL is prone to getting stuck in local optima. However operational space control could mitigate this problem by specifying high-level goals in task-space. \cite{khatib1987,schaal2007,schaal2008,schaal2011}. This corresponds to shaping the control actions so that policies only search the space where a ``good" solution exists. 

We seek to answer the following three questions:
\begin{enumerate}
    \item Will it help if control actions stem from an operational space controller? Since tool space forces provide the most straightforward and explicit information of such force-based tasks, can local trajectory optimizer with Markovian properties benefit from them?
    \item Can adaptive impedance behavior be learned by our methods?
    \item Can we learn a policy with generalization capabilities to local variations by explicitly considering force/torque measurements?
\end{enumerate}

   \begin{figure*}[t]
       \centering
   \includegraphics[width=1\textwidth,keepaspectratio]{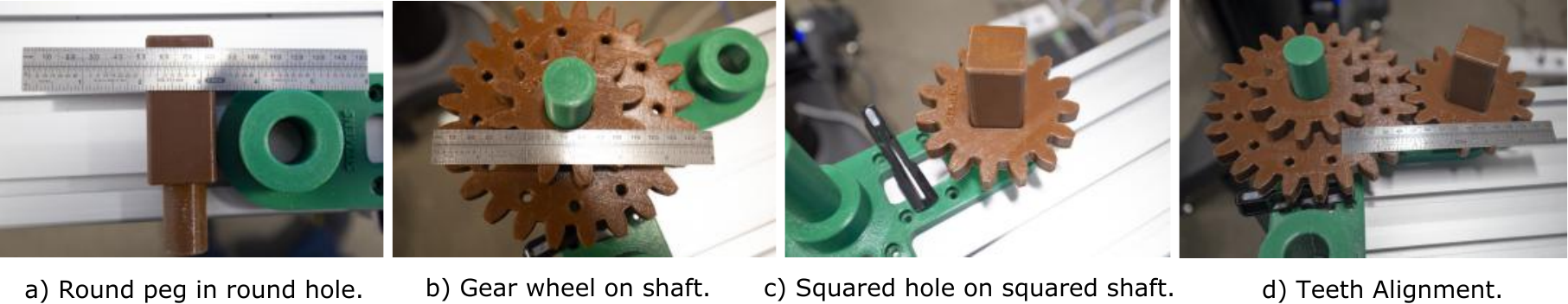}
      \caption{Four tasks that represent different assembly challenges. Each task requires a flexible control policy that needs to consider contacts and friction. Sub-figure a) to d) represents task 1 to 4 respectively.}
       \label{assemble}
    \end{figure*}

The contributions of this paper are answers to these questions. First, we find that local trajectory optimization can significantly benefit by using operation space controller for control actions.
Second, we show adaptive compliant behavior can be acquired autonomously through interactions.
Third, we propose a method to incorporate force/torque sensor data into global policy parameterized by neural networks. 

%% file: sections/02_problem.tex
\section{PROBLEM STATEMENT AND RELATED WORK}

\paragraph{Problem Statement}

Consider the task of assembling the gear set shown in Fig.~\ref{cover_photo}. 
The gear model was introduced by Siemens Corporation as a benchmark task for robotic assembly\footnote{\url{http://www.usa.siemens.com/robot-learning}}. 
The overall assembly task consists of four sequential steps, which are illustrated in Fig.\ref{assemble}: first the robot needs to insert a cylindrical peg into its matching hole; then the large brown gear should be inserted through the gear shaft; then the small brown gear with the squared hole should be assembled; lastly the gear wheels need to be matched by aligning the corresponding gear teeth. In general the tolerances are tight. For example, step two requires tolerances tighter than 0.1 mm, which is beyond most deployed industrial robots' accuracy today. Additionally, in step two, the peg can freely rotate at contact, the gear must be precisely oriented to match the squared peg; in step three, the small brown gear must be rotated by the large brown gear properly so that they can align with each other. This poses additional challenges: since none of these pegs or gears are fixed during assembly, this added uncertainty makes assembly even more difficult.

\paragraph{Related Work}
Recent advances in RL have gained great success in solving a variety of problems from playing video games \cite{mnih2013playing,mnih2016asynchronous,lillicrap2015continuous} to robotic locomotion \cite{abbeel2008nips,jianlan_icra18,schulman2015trust,sergey_nips2014,zhang2016learning}, manipulation \cite{sergey_icra15,chebotar2017path,peters2010relative,fu2016one,levine2016end,aviv2017icra,marcIJRR,deepmpc,jianlan_iros18,residualrl}. 
Reinforcement learning can be distinguished in model-based methods and model-free methods \cite{sutton1998reinforcement,kober2013reinforcement,deisenroth2013survey}.
While model-based policy search is computationally more expensive than model-free methods, it requires less data to solve a task.
Recent progress in the area of Deep Neural Networks suggests deploying them for parametrizing policies and other functions in RL methods \cite{levine2016end, levine2013guided,schulman2015trust}.
This is often referred to as Deep Reinforcement Learning (DRL).

A recently developed model-based reinforcement learning algorithm called guided policy search (GPS) provided new insights into training end-to-end policy for solving contact-rich manipulation problems \cite{sergey_nips2014, levine2013guided,MDGPS,levine2016end,pilqr_icml17}; however; this method is not suitable for this high-precision setting because it has no means of avoiding local optima by its formulation.
There are also approaches tackling this problem by explicitly modeling contact dynamics \cite{billard2018,billardtro,cio,igo2015iros,igo2015nips}
Inoue et al. \cite{inoue2017deep} use LSTM to learn two separate policies for finding and inserting a peg into a hole; however, their methods require several pre-defined heuristics, and also the action space is discrete.

Thomas et al. \cite{aviv2018icra} combine RL with a motion planner to shape state cost in high-precision settings. This method essentially learns a trajectory following torque controller, and assumes access to a trajectory planner that could roughly avoid local optima. They also encode such planned reference into a neural network with attention mechanism, they show good generalization results in simulation.

%% file: sections/03_prelims.tex
\section{PRELIMINARIES}
We consider all tasks here that can be described as moving already-grasped objects to their goal position. This is the most common setting in today's manufacturing. The success of such tasks can be measured as minimizing the distance between objects and their goal positions. We make no particular assumptions about encountered dynamics during tasks especially contacts. These need to be learned by the robot from various interaction with its environment. Let $\bx_t$ and $\bu_t$ denote robot states and actions respectively ; $\ell(\bx_t,\bu_t)$ be the cost function related to the task, $T$ be the time horizon of a task.  Our problem can be formulated as
\begin{align*}
    &\min_{\bu_1,\bu_2 ...\bu_T}\sum_{t=1}^{T}\ell(\bx_t, \bu_t) \\
    &s.t. \quad \bx_{t+1} = f(\bx_t,\bu_t) \,\, t=1,2...T-1,
\end{align*}
where f governs (unknown) system dynamics, $\bx$, $\bu$ can also be subject to other algebraic constraints.

We consider our control action $\bu$ to be operational force controller $\mathcal{F}_{tip} = [F_x,F_y,F_z,M_x,M_y,M_z]$. They represent desired force/torque or impedance in operational space, our goal is to optimize them through reinforcement learning. 

%% file: sections/04_rl_force.tex
\section{Reinforcement Learning with Force Control}\label{RLFORCECONTROL}
In this section, we first introduce operational space force control, and then move towards hybrid motion/force controller for more stable behaviors. We pick a particular model-based RL algorithm, iterative Linear-Quadratic-Gaussian (iLQG) for combining with these controllers, because it has been shown to be sample efficient. We then propose a neural network architecture that explicitly considers force information for better generalization purpose. 

We explain our intuition for combining an operational force controller with RL using Fig.\ref{assemble}(b). One successful strategy for inserting the gear with such high accuracy is to constrain the motion and forces somehow. For instance, if gear and the stand are in contact, we cannot move the gear downwards unless they are aligned. This is a natural constraint due to the rigid nature of environment. It is obvious that the task simplifies, if we constrain the motion to planar motions during the hole-searching phase; tilting would only help with fine adjustment when the gear is being inserted with high friction, this is an artificial constraint imposed by humans. Careful combinations of natural constraints and artificial constraints are essential to generate ``task descriptions" in high-precision settings considered in this paper. Indeed, these can be regarded as Pfaffian constraints consisting of holonomic and nonholonomic components. Our method could be thought as implicitly generating such constraints. 

Besides improving insertion accuracy we also learn a policy that is robust to local variations. We propose a neural network architecture as seen in Fig. \ref{nn}, where force/torque measurements of the current time step are explicitly considered for control action derivation. 
\subsection{Operational Space Motion/Force Controller}\label{op_space_force}
Let $\mathcal{F}_{tip}$ be the desired wrench on the end-effector, we can then express the control law in joint space as \cite{modernrobotics}
\begin{equation}\label{force_eq}
    M(q)\ddot{q} + c(q,\dot{q})\dot{q} + g(q) + J^{T}(q)\bF_{tip} = \tau,
\end{equation}
where $q$ represents joint angles in generalized coordinates, $M(q)$ is the inertia, $c(q,\dot{q})$ is the Coriolis matrix, $g(q)$ are gravitational forces, $J(q)$ is the Jacobian, $\tau$ is the torque vector applied to manipulators' joints.
In many force control tasks, robots move slowly, hence we ignore acceleration and velocity terms in Eq. \ref{force_eq}.
For a 7-DOF Sawyer manipulator arm that we consider in this paper, we can also project torques to its non-empty nullspace. Denoting the nullspace torque vector as $\tau_{null}$, joint space control law is:
\begin{equation}\label{force_eq_2}
    \tau = g(q) + J^{T}(q)\bF_{tip} + [I - J^{T}(q)J^{T\dagger}(q)]\tau_{null},
\end{equation}
where $J^{T\dagger}(q)$ is the pseudo-inverse of $J^{T}(q)$. The control law in Eq.\ref{force_eq_2} is appealing and simple, but it would generate undesirable and dangerous motion without enough resistance provided by the environment. In our experimental setup, we do not assume in-contact situations of objects being assembled, there is a relative open free-space that the robot needs to move through; directly applying Eq.\ref{force_eq_2} would result in continuous acceleration. To mitigate this issue, in all our experiments, we wrap a position loop with small gains around the controller in Eq.\ref{force_eq_2}:
\begin{align}\label{weak_hybrid}
      \tau &=  \Sigma_1 [K_{qp}(q - q^{\ast}) + K_{qd}(\dot{q} - \dot{q}^{\ast})] + \Sigma_2 J^{T}(q)\bF_{tip} \\
      &+ [I - J^{T}(q)J^{T\dagger}(q)]\tau_{null} + g(q) ,
\end{align}
where $K_{qp}$ and $K_{qd}$ are diagonal gain matrices with small entries, $q$ and $\dot{q}$ are current joint positions and velocities, $q^{\ast}$ and $\dot{q}^{\ast}$ are the desired ones obtained via inverse kinematics from end-effector pose. $\Sigma_1$ and $\Sigma_2$ are diagonal matrices to weight motion and force portions, respectively. The resulting hybrid controller also achieves adaptive impedance behavior implicitly: $\bF_{tip}$ is a time-varying linear-Gaussian controller conditional on robot configuration, which is detailed in Sec. \ref{ilgq}. This learned piecewise linear controller will ideally yield high-impedance when moving in free-space, and high-admittance whenever in contact; thus implicitly scaling motion-to-force ratio in aforementioned controller.
Aforementioned $\bF_{tip}$ will be calculated by an RL controller.

\subsection{Iterative Linear-Quadratic-Gaussian Controller}\label{ilgq}
The specific model-based reinforcement learning algorithm that we consider here is iterative Linear-Quadratic-Gaussian (iLQG). It is sample efficient and convenient second-order methods are available to solve it quickly \cite{ilqg}. Let $\omega= \{\bx_1,\bu_1,\,...\,,\bx_T,\bu_T\}$ denote a trajectory, such that $p(\omega) = p(\bx_1)\prod\limits_{t=1}^{T}p(\bx_{t+1}|\bx_{t},\bu_{t})p(\bu_t|\bx_t )$,  $\ell(\omega) = \sum_{t=1}^{T}\ell(\bx_t, \bu_t)$ denotes the cost along a single trajectory $\omega$; where $\bx$ typically consists of joint angles, end-effctor pose and their time derivatives.  We wish to minimize this cost; the goal is to minimize the expectation $E_{p(\omega)}[\ell(\omega)]$ over trajectory $\omega$ by iteratively optimizing linear-Gaussian controllers and re-fitting linear-Gaussian dynamics. 
This algorithm iteratively linearizes the dynamics around the current nominal trajectory, constructs a quadratic approximation of the cost, computes the optimal actions with respect to this approximation of the dynamics and cost by dynamic programming, and forward runs resulting actions to obtain a new nominal trajectory. 
We adopt a slightly different version of iLQG proposed in \cite{ZiebartMaxEnt,sergey_nips2014}. An additional entropy term is added into cost function such that $\tilde{\ell}(\bx_t,\bu_t) = \ell(\bx_t,\bu_t) - \mathcal{H}(p(\bu_t|\bx_t))$ to encourage exploration. 
It can be shown that the optimal control law to this problem is $p(\bu_t|\bx_t) = \mathcal{N}(\bK_t \bx_t + \bk_t, \mathbf{C}_t)$, and $\mathbf{C}_t = Q^{-1}_{\bu,\bu t}$, where $Q$ is cost to go, $\mathbf{K}_t = -Q^{-1}_{\mathbf{u,u}t} Q_{\mathbf{u,x}t}$ and $\mathbf{k}_t = -Q^{-1}_{\mathbf{u,u}t}Q_{\mathbf{u}t}$, subscripts denote ordered derivatives at time $t$ \cite{ZiebartMaxEnt}.

\subsection{Interpretation as Learning Pfaffian Constraints}\label{hybrid_motion_force}
 Our method could be regarded as generating Pfaffian constraints for each task. At contact, we can formulate holonomic and nonholonomic constraints enforced by the rigid environment as Pfaffian constraints:
\begin{equation}\label{constraint_force}
    A(q)\mathcal{V} = 0,
\end{equation}
where $\mathcal{V}$ is the operational space twist in \textit{SE}(3) that $\mathcal{V} \in \mathbb{R}^6$, $A(q) \in \mathbb{R}^{k \times 6}$, $k$ is the number of natural constraints. In motion control part, we can write down operational space dynamics of the robot as:
\begin{equation}\label{operational_dynamics}
    \mathcal{F} = \Lambda(q)\dot{\mathcal{V}} + \eta(q,\mathcal{V}), 
\end{equation}
where $\Lambda(q)=J^{-T}(q)M(q)J^{-1}(q)$, 
$\eta(q,\mathcal{V}) = J^{-T}(q)c(q,J^{-1}\mathcal{V}) - \Lambda(q)\dot{J(q)}J^{-1}(q)\mathcal{V}$. This is essentially the same motion dynamics without $\bF_{tip}$ expressed in Eq.\ref{force_eq}, but calculated in operational space. 

By combining Eq.\ref{constraint_force} and Eq.\ref{operational_dynamics}, we can express constrained dynamics for this hybrid motion/force controller as
\begin{equation}\label{hybrid_force_ctrl}
    \mathcal{F} = \Lambda(q)\dot{\mathcal{V}} + \eta(q,\mathcal{V}) + \underbrace{A^{T}(q)\lambda}_{\bF_{tip}},
\end{equation}
where $\lambda \in \mathbb{R}^k$; and in this case, requested wrenches $\bF_{tip}$ must lie in the column space of $A^{T}(q)$. 
In general, constraints $A(q)$ come from the environment the robot is interacting with. Abstracting this type of constraint for each individual manipulation task can be time-consuming and prone to errors. Instead, if we let $\bF_{tip}$ be learned by RL through continuous interactions; we can roughly regard this process as iteratively improving $A(q)$ to generate increasingly accurate description of tasks, which is a key ingredient in high-precision settings.

\subsection{Training Neural Network controller using MDGPS}\label{nn_train}
\begin{figure}[thpb]
       \centering
   \includegraphics[width=0.45\textwidth,keepaspectratio]{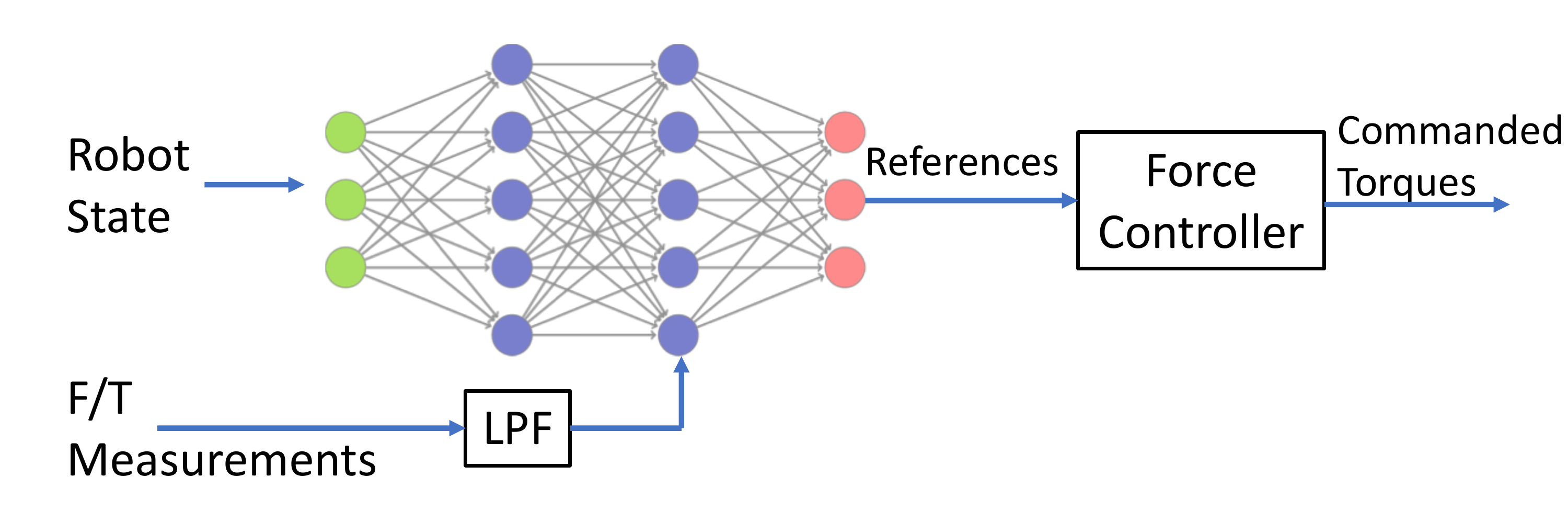}
      \caption{Neural Network Architecture}
       \label{nn}
    \end{figure}

We introduce a novel neural network architecture to process noisy force/torque readings from wrist sensors or other sources, which provide measurements in tool space. The neural network is shown in Fig. \ref{nn}. Force/torque information is filtered with a low-pass filter (LFP), then concatenated in the second last layer of the neural network. Intuitively, we would like to provide most direct haptics information to the neural network as principle features; also avoiding the neural network to establish unreasonable correspondence between external force/torque readings and robot internal states. Also, the robot needs to move in free space before it is in contact; and the learned policy should be robot-configuration-dependent rather than force-dependent in free space. Since F/T readings are noisy, the learned policy dependent on robot state and coupled F/T generates random motions, if we directly treat F/T as an input to the first layer of the neural network. 

For training this neural network, we adopt the mirror descent guided policy search (MDGPS) algorithm in \cite{MDGPS}. Denote the neural network parameterized by $\theta$ as $\pi_{\theta}(\bu_t|\bx_t)$, the goal is to minimize the KL-divergence between linear-Gaussian controller learned via iLQG:
\begin{equation*}
    \min \mathbb{D}_{KL}(\pi_{\theta}(\bu_t|\bx_t)||p(\bu_t|\bx_t))
\quad \forall \bx_t, \bu_t,t,
\end{equation*}
this can be implemented by supervised learning. These guided policy search methods also have some mechanism to enforce agreement between distributions of local policy and global policy by adding an additional KL-divergence cost \cite{MDGPS,levine2013guided,levine2016end}.

We summarize our method in Alg.\ref{overall}.

\begin{algorithm}
	\caption{Force-based RL controllers}
	\begin{algorithmic}[1]
		\FOR{iteration $k \in \{1,..., K\}$}
		\STATE{Train local RL controller using iLQG, where $\bu_t$ is set as operational space force controller}
		\STATE{Project calculated operational control to joint torque using Eq.\ref{weak_hybrid}  }
		\STATE{Train neural network controller using MDGPS\cite{sergey_nips2014}}

		\ENDFOR
	\end{algorithmic}
	\label{overall}
\end{algorithm}

%% file: sections/05_experiments.tex
\section{EXPERIMENTS}
In this section we answer the following questions. (1) How does the proposed iLQG with force control perform? Is it actively exploiting contact constraint dynamics as we hypothesized? How does it compare to its ablations where force information is integrated differently? (2) How does the proposed neural network architecture improve local generalization? How does it compare to its ablations? 

\subsection{Experimental Setup Details}
We evaluate our methods on four assembly tasks, which are shown in Fig.\ref{assemble}.
We use a Rethink Robotics Sawyer robot. Sawyer offers an interface to query its wrist force/torque measurement, the noise levels (estimated standard deviations) for $F_x, F_y$ are 2.0\,N; $F_z$ is 0.5\,N; $M_x,M_y$ are 0.5\,Nm; and $M_z$ is 0.1\,Nm.  Sawyer is commanded via ROS at 20 Hz. During training, we take four roll-outs per iteration. Typically, it takes three iterations to achieve successful behaviors, five iterations for convergence. We define a plane by three points in end-effector space, the cost function is a weighted mixture of the $\ell_1$ and $\ell_2$ norms of the differences between the current plane and the target plane as specified by the three aforomentioned points.

\subsection{Assembly Performance Results}
We compare our method of Sec. \ref{ilgq} with the following baselines: 
\begin{itemize}
    \item \textbf{Kinematics Only:} For task 1 and task 2, we only specify target poses; for the more difficult task 3 and task 4, we also introduce several way-points. Note that for task 1 and 2, we should get the same result every single time since robot kinematic controller is deterministic as well as these tasks. But for task 3 and 2, the peg and gear can move freely, so it is hard to specify the desired trajectory.
    \item \textbf{iLQG with torque control:} This is the main baseline for comparison. The control actions from iLQG are directly the seven joint torques. For comparison, we use the same cost function as in our method, i.e., sparsely-defined target end-effector pose, no intermediate way-points are introduced.
    \item \textbf{iLQG with torque control, augmented state space:} We augment the state space with the F/T vector such that $\tilde{\bx_t} = [\bx_t, f_t]$, where $f_t$ are F/T measurements. We apply direct torque control. The purpose of this is to verify if other formulation other than what we proposed could also actively use this additional information. 
    \item \textbf{Our Method:} We refer to the operational space controller in Section \ref{RLFORCECONTROL} with iLQG.
    \item \textbf{Our Method with augmented state space:} Additional to operational space controller, we augment the state space to $\tilde{\bx_t} = [\bx_t, f_t]$. This experiment is for verifying if our method can be further improved.
\end{itemize}
A success is considered if an object is being assembled to a desired pose with defined tolerance. We report success rates for each individual task separately, because we train an individual policy for each task. However, it would be straightforward to report overall success rate by multiplying individual success rates together since policies are trained independently. We execute learned policies after training to calculate the success rates. Table \ref{table1} presents aforementioned success rates for four different tasks.
\begin{figure}[tbph]
       \centering
   \includegraphics[width=0.48\textwidth,keepaspectratio]{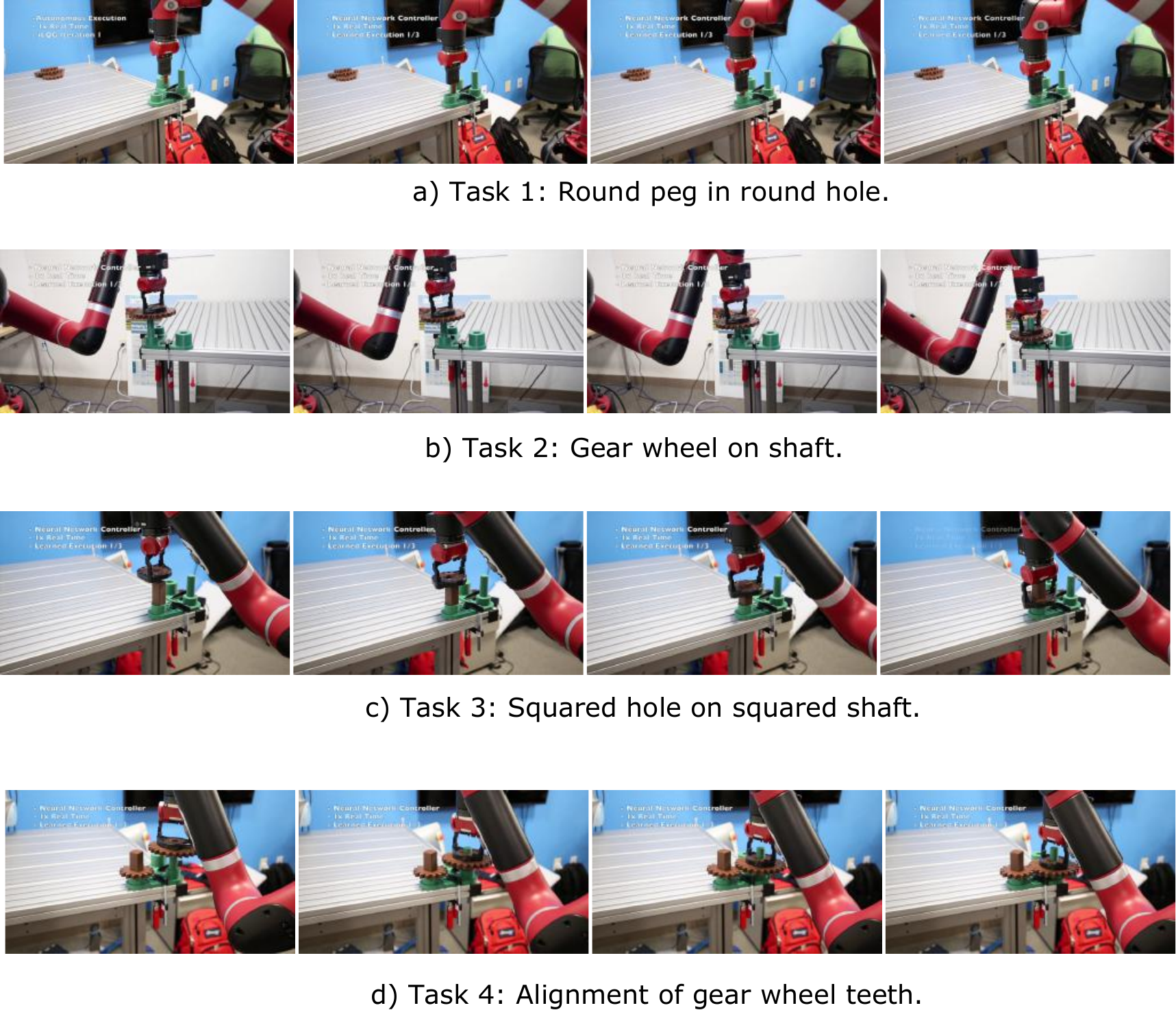}
      \caption{Snapshots of experimental runs for the four studied assembly tasks. }
       \label{experiments}
    \end{figure}
\begin{table}[h!]
\centering
\caption{Comparison of success rates for different tasks. Baseline 1 refers to kinematics only; baseline 2 refers to iLQG with direct torque control; baseline 3 refers to iLQG with direct torque control, augmented state space, our method w/ augmented refers augmented state space in our method.}
\begin{tabular}{|c|| c | c| c| c|} 
 \hline
  & Task 1 & Task 2 & Task 3 & Task 4 \\ [1ex] 
 \hline
  baseline 1 & 0/5 & 0/5 & 0/5 & 0/5 \\ [1ex] 
 \hline
 baseline 2& 1/5 & 0/5 & 0/5 & 0/5 \\
 [1ex] 
 \hline
  baseline 3& 0/5 & 0/5 & 0/5 & 0/5 \\
 [1ex] 
 \hline
  our method& \textbf{5/5} & \textbf{5/5} & 2/5 & \textbf{4/5} \\
 [1ex] 
 \hline
  our method w/ augmented & 5/5 & 5/5 & \textbf{3/5} & 3/5  \\
 [1ex] 
 \hline
\end{tabular}
\label{table1}
\end{table}

We interpret these results several fold: (1) kinematics baseline fails consecutively, this confirms the required accuracy and complexity for the gear set; (2) an iLQG with torque control, but without extensive cost shaping fails; the single success we observed is due to Gaussian noise in the controller, which generated some lucky motion to insert, and it is on the easiest task. (3) we did not find reliable improvement by augmenting state space with F/T information. Since F/T signals are not Markovian, fitting a time-correlated dynamics model to them does not produce meaningful information.

We made several interesting observations during the experiments.
 During task one, the robot moves quickly in free-space to reach in-contact status, then it reduces its speed to slowly probe around, trying to "feel" the surface; once it has a level of confidence of the hole's position, it becomes aggressive towards the goal it predicted, resulting in quick motions followed by a large downward force to complete insertion. The most interesting experiment is task 3, where the added uncertainty comes from a rotating peg. The robot first brings the small gear in contact with the peg, while applying a downward force so that small gear would not fall into free-space again; but this amount of downward force also allows room for applying additional rotating torque to the peg and gear aligning them with each other roughly; then the downward force increases to try insertion, if not successful, downward forces decrease but small horizontal force are also observed to fine-tune poses, this procedure iterates until the peg is fully inserted. This kind of behavior roughly aligns with humans' heuristics when facing such tasks. These behaviors can be found in the supplemental video\footnote{\url{https://www.youtube.com/watch?v=kqA1tlPaT8E}}.

\begin{figure}[thpb]
       \centering
   \includegraphics[width=0.5\textwidth,keepaspectratio]{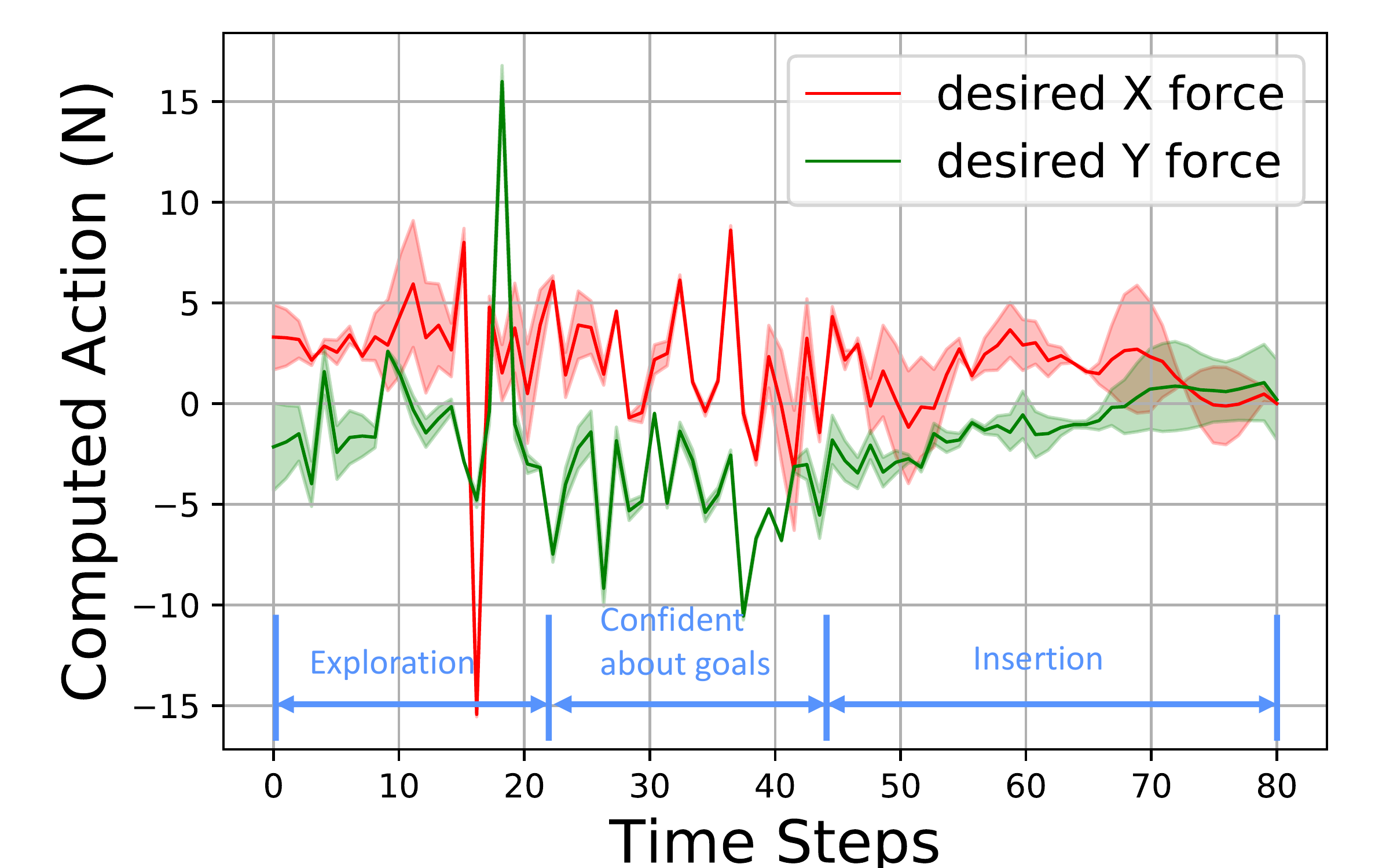}
      \caption{Action computed by learned policy during one successful insertion. Solid line shows computed action mean, and error bar for computed variance.}
       \label{policy_action}
    \end{figure}

\begin{figure}[thpb]
       \centering
   \includegraphics[width=0.5\textwidth,keepaspectratio]{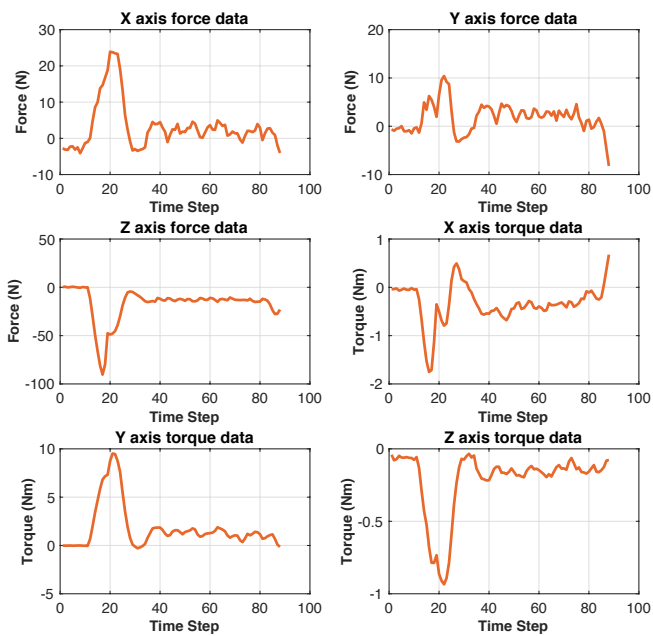}
      \caption{Six degree of freedom force torque measurements from a successful Task 2 insertion. }
       \label{ftdata}
    \end{figure}

Fig.\ref{policy_action} shows computed actions (only desired forces in x-directions and y-directions are plotted) for task 2 during one successful insertion. It is interesting to observe how the variance computed by the policy changes over time. Initially, there is a certain level of variance for exploration to search for the target position; once the policy is confident about the goal, the variance reduces dramatically, the robot aggressively moves the object towards the goal; finally during the insertion phase, a certain level of noise is again injected for fine-tuning the gear's pose to overcome friction. This force-based insertion pattern is automatically discovered  through interactions by the algorithm, and matches a human's intuition on such tasks well. Fig.\ref{ftdata} presents F/T measurements during a successful insertion. We can observe peaks both in force and torque data, indicating some critical phases, e.g., contact. This motivates explicit use of F/T measurements, because of its informative nature.

\subsection{Generalization Results}
We only train the neural network controller based on a single instance of the local linear-Gaussian controller. Hence, our purpose is not to see if the trained neural network controller can interpolate between multiple local controllers as in original guided policy search methods \cite{levine2016end}. Instead, we are interested in examining if the proposed architecture can effectively use F/T information and adapt to environment variations. We compare our proposed neural network against two baselines: one is to directly input F/T information into the first layer of the neural network; second is the iLQG controller that the neural network controller is trained on. We again only consider task 2 in these generalization experiments. We design our experiments as following: we train these three policies towards with the goal fixed, i.e., base does not move; after all policies are trained, we move base to slightly different positions but policies will be kept unchanged; then we count success rates for these variations of different base positions. The intuition is that even base is moved, the proposed neural network controller should still be able to find the hole if F/T information is actively used, since variations in base positions bear same force pattern in terms of peg-hole insertions. The comparison with iLQG baseline is due to the fact that slight variations in the goal position could also result in low cost in LQR. We want to distinguish between this and active F/T information. 

We test these three methods in three different settings where the base is moved 1cm, 2cm, 5cm respectively. Success rates are reported in Table \ref{generalize}

\begin{table}[h!]
\centering
\caption{Comparison of success rate in generalization capability }
\begin{tabular}{|c ||c| c| c|} 
 \hline
  & 1cm & 2cm & 5cm \\ [1ex] 
 \hline
 iLQG & 8/10 & 5/10 & \textbf{6/10} \\ [1ex] 
 \hline
 our method & \textbf{9/10} & \textbf{8/10} & \textbf{6/10} \\
 [1ex] 
 \hline
 F/T input to first layer & 0/10 & 0/10 & 0/10 \\
 [1ex] 
 \hline
\end{tabular}
\label{generalize}
\end{table}
For neural networks that input F/T data to their first layer, the training was often aborted due to large KL-divergence between iLQG and neural network; for few scenarios we could successfully train a policy, it often generates random, undesirable motions in free-space before insertion; it behaved poorly on three experiments. This validates our hypothesis for not establishing correspondence between external F/T readings and robot internal states.

For the comparison of the proposed neural network controller against iLQG controller, we found that neural network controller produced slightly better results; but at the mean time, the variance of the result was also high. The neural network controller did not outperform iLQG consistently. Getting a more accurate six axes F/T controller might mitigate this issue.

%% file: sections/06_conclusion.tex
\section{CONCLUSIONS AND FUTURE WORK}
In this paper we combine RL with an operational space force controller to solve the problem of high-precision assembly. We show that RL essentially automates the generation of Pfaffian constraints in operational space constrained dynamics, which we regard as a crucial ingredient for high-precision tasks. We specifically exploited one of the model-based RL algorithm, iLQG, compared with several ablations, results show that our method performs best in this high-precision settings. We also introduced a neural network architecture that explicitly considers force/torque information in decision making process, which leads to a better result of generalization. 

 One future direction is to add raw vision and tactile inputs to the architecture in Figure \ref{nn}, thus becoming an end-to-end multi-modal neural network. It would be interesting to see if such learned policy can succeed from arbitrary starting positions in free space, and tactile sensing could further improve policy performance.
 Another interesting direction is to explicitly model contact, and encode such information as priors for a more structured Pfaffian constraint matrix, this would further reduce sample complexity and serve as a general primitive for policy transfer.

%% file: sections/acknowledgement.tex
\section*{ACKNOWLEDGEMENT}
This work is partially funded by Siemens. Authors would like to thank Tobias Johannink for generous help on setting up the experiments.